\documentclass[final,3p,times,twocolumn]{elsarticle}

\usepackage[numbers]{natbib}
\usepackage{amsmath,amsfonts}
\usepackage{algorithmic}
\usepackage{algorithm}
\usepackage{array}
\usepackage{caption}
\usepackage{subcaption}
\usepackage{enumerate}
\usepackage{textcomp}
\usepackage{stfloats}
\usepackage{url}
\usepackage{verbatim}
\usepackage{graphicx}
\usepackage{cite}

\usepackage{epsfig} 
\usepackage{amssymb}  
\usepackage{multirow}
\usepackage[table]{xcolor}
\usepackage{import}
\usepackage{booktabs}
\usepackage{tabularx}
\usepackage{gensymb} 
\usepackage{bm}
\usepackage{siunitx}
\usepackage{csquotes}
\usepackage{acro}
\DeclareAcronym{com}{ 
    short = {CoM}, 
    long  = {Center of Mass},
    first-style = short-long,
}
\DeclareAcronym{cog}{ 
    short = {CoG}, 
    long  = {Center of Geometry},
    first-style = short-long,
}
\DeclareAcronym{zmp}{ 
    short = {ZMP}, 
    long  = {Zero Momentum Point},
    first-style = short-long,
}
\DeclareAcronym{wrt}{ 
    short = {w.r.t.}, 
    long  = {with respect to},
    first-style = short-long,
}
\DeclareAcronym{dof}{ 
    short = {DoF}, 
    long  = {Degree of Freedom},
    first-style = short-long,
}
\newcommand\new[1]{\textcolor{black}{#1}}

\journal{Robotics and Autonomous Systems}

\begin{document}

\begin{frontmatter}



\title{Enhancing Sliding Performance with Aerial Robots: Analysis and Solutions for Non-Actuated Multi-Wheel Configurations\tnoteref{star}} 
\tnotetext[star]{This work has been supported by the European Unions Horizon 2020 Research and Innovation Programme AERO-TRAIN under Grant Agreement No. 953454.}

\author[1]{Tong Hui\corref{cor1}} 
\ead{tonhu@dtu.dk}
\cortext[cor1]{Corresponding author}
\affiliation[1]{organization={Technical University of Denmark},
            city={Kongens Lyngby},
            postcode={2800}, 
            country={Denmark}}
\affiliation[3]{organization={Autonomous System Lab, ETH},
            addressline={Zurich}, 
            country={Switzerland}}
\affiliation[2]{organization={Univ Rennes, CNRS, Inria, IRISA},
            addressline={F-35000 Rennes}, 
            country={France}}
\author[3]{Jefferson Ghielmini}
\ead{jghielmini@mavt.ethz.ch}
\author[1]{Dimitrios Papageorgiou}
\ead{dimpa@dtu.dk}
\author[3]{Roland Siegwart}
\ead{rolandsi@ethz.ch}
\author[2]{Marco Tognon}
\ead{marco.tognon@inria.fr}
\author[1]{Matteo Fumagalli}
\ead{mafum@dtu.dk}

\begin{abstract}
Sliding tasks performed by aerial robots are valuable for inspection and simple maintenance tasks at height, such as non-destructive testing and painting. Although various end-effector designs have been used for such tasks, non-actuated wheel configurations are more frequently applied thanks to their rolling capability for sliding motion, mechanical simplicity, and lightweight design. Moreover, a non-actuated \textit{multi-wheel} (more than one wheel) configuration in the end-effector design allows the placement of additional equipment e.g., sensors and tools in the center of the end-effector tip for applications. However, there is still a lack of
studies on crucial contact conditions during sliding using aerial robots with such an end-effector design. In this article, we investigate the key challenges associated with sliding operations using aerial robots equipped with multiple non-actuated wheels through in-depth analysis grounded in physical experiments. The experimental data is used to create a simulator that closely captures real-world conditions. We propose solutions from both mechanical design and control perspectives to improve the sliding performance of aerial robots. From a mechanical standpoint, design guidelines are derived from experimental data. From a control perspective, we introduce a novel pressure-sensing-based control framework that ensures reliable task execution, even during sliding maneuvers. The effectiveness and robustness of the proposed approaches are then validated and compared using the built simulator, particularly in high-risk scenarios.
\end{abstract}




\end{frontmatter}



\section{Introduction}
With the growing interest in utilizing aerial robots for industrial applications, there has been a notable increase in research that pertains to aerial systems engaging in physical interactions with the environment. The aerial robotics community is actively researching diverse physical interaction tasks \citep{anibal2022}, with the primary goal of broadening the range of potential industrial applications. Sliding tasks are of particular interest due to their applicability in several use cases, e.g., continuous scanning for Non-Destructive Testing (NDT) \citep{tognon2019,truj2019,zhao2019,antonio2019,nava2020,watson2022,hui2023} such as ultrasonic testing, layouting \citep{chris2022,karen2019}, cleaning and grinding \citep{woper2018}. During sliding with aerial robots, the physical contact between the end-effector tip and the work surface is typically preserved by exerting a pushing force on the work surface. Tasks involving the aforementioned interactions are commonly known as the push-and-slide tasks. Often a rather large pushing force is required to have stable contact \citep{hui2023} which significantly increases the energy consumption of the platform.  
\begin{figure}[!t]
\centering
\begin{subfigure}{0.45\columnwidth}
    \centering
    \includegraphics[trim={0.1cm 0.1cm 0.1cm 0.1cm},clip,width=\columnwidth]{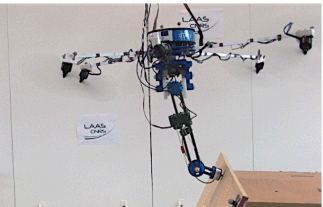}
    \caption{}
    \label{fig:nonwheel}
\end{subfigure}
\begin{subfigure}{0.45\columnwidth}
    \centering
    \includegraphics[trim={0.1cm 0.1cm 0.1cm 0.7cm},clip,width=\columnwidth]{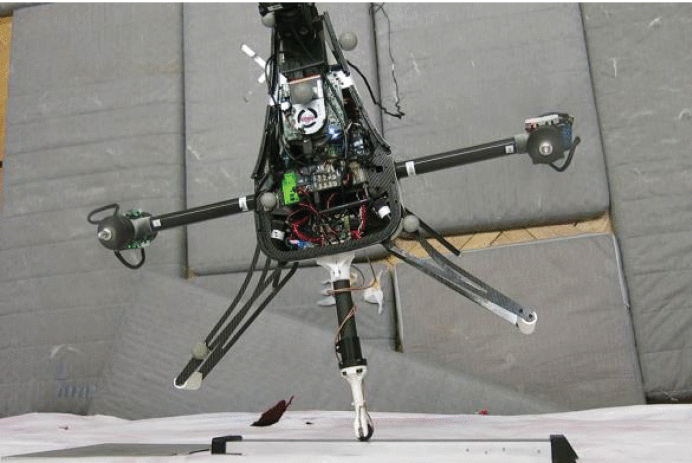}
    \caption{}
    \label{fig:1wheel}
\end{subfigure}
\begin{subfigure}{0.45\columnwidth}
    \centering
    \includegraphics[trim={0.1cm 0.1cm 0.1cm 0.1cm},clip,width=\columnwidth]{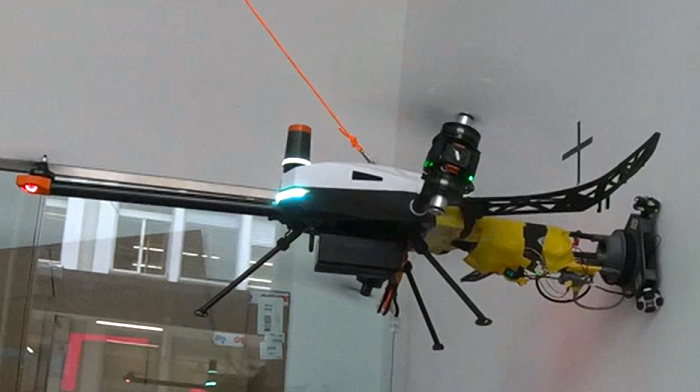}
    \caption{}
    \label{fig:3wheel}
\end{subfigure}
\begin{subfigure}{0.45\columnwidth}
    \centering
    \includegraphics[trim={0.1cm 0.1cm 0.1cm 0.5cm},clip,width=\columnwidth]{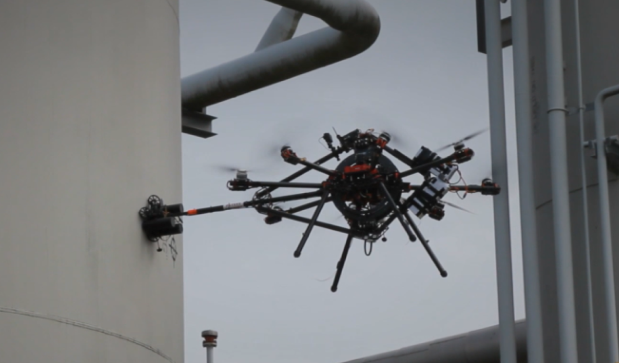}
    \caption{}
    \label{fig:4wheel}
\end{subfigure}
\caption{Examples of different end-effector designs for sliding tasks with aerial robots. (a) Non-wheeled design \citep{nava2020}. (b) One-wheeled design \citep{watson2022}. (c) Actuated three-wheeled design \citep{hui2023}. (d) Non-actuated four-wheeled design \citep{truj2019}.}
\label{fig:intro}
\end{figure}
\subsection{End-effector designs for sliding}
The physical contact for different interaction tasks between the robot and the environment is in general achieved through specialized end-effectors. A range of different types of end-effectors with variable contact points has been studied in current literature for sliding tasks involving non-wheeled and wheeled designs, see Fig.~\ref{fig:intro}. In \citep{antonio2019,nava2020,zhang2022learning,peric2021}, non-wheeled end-effectors with one contact point are tested. 
With these end-effectors, the system has to overcome the static friction to start a sliding motion as in Fig.~\ref{fig:nonwheel}. This often causes a sudden acceleration in the system due to the switch from static to dynamic friction. Moreover, the static and dynamic friction coefficients vary on different work surfaces \citep{zhang2022}. Therefore, non-wheeled end-effectors are not the best for sliding tasks considering diverse testing conditions. In \citep{karen2019,watson2022}, a one-wheeled end-effector is used which also has one contact point but is more widely used for sliding motions thanks to the wheel's rolling ability, see Fig.~\ref{fig:1wheel}. However, wheeled end-effectors with only one contact point hinder the placement of additional equipment e.g., sensors and tools, which limits the application range of the robot.
In this paper, we refer to systems equipped with more than one wheel as \textit{multi-wheeled systems}, e.g., \textit{three-wheeled systems}, \textit{four-wheeled systems}. In \citep{woper2018,zhao2019,truj2019,chris2022,hui2023}, three-wheeled and four-wheeled end-effectors with multiple contact points are implemented which allow for the mounting of tools and sensors in the middle of the end-effector tip. While a multi-wheeled end-effector is sliding on the work surface, one of the main objectives is to ensure stable contact between all wheels of the end-effector and the work surface \citep{anibal2022}.

For sliding, we define the use of actuated wheels as \textit{active sliding}, and the use of non-actuated wheels as \textit{passive sliding}. \citet{woper2018}, \citet{chris2022}, and \citet{hui2023} demonstrate multi-wheeled active sliding with underactuated as well as fully-actuated aerial vehicles on flat surfaces, see Fig.~\ref{fig:3wheel}. Active sliding imposes relatively simple control requirements on the aerial system to maintain the body orientation and the pushing force, while the sliding motion is driven directly by the wheel motors \citep{hui2023}. Despite the simple control requirements of active sliding, the addition of wheel motors often leads to bulky end-effector designs and additional weight. On the other hand, end-effectors with non-actuated wheels feature a simple mechanical design and are lightweight. \citet{zhao2019} presents passive sliding with a four-wheeled end-effector in simulations. \citet{truj2019} showcased pipe inspection using an aerial vehicle equipped with a 6-\ac{dof} arm and four non-actuated wheels as in Fig.~\ref{fig:4wheel}. Yet, there is still a lack of study on the passive sliding problem with the simplest system design where multiple non-actuated wheels are rigidly attached to the aerial vehicle without additional \ac{dof}.

\subsection{Aerial vehicles for multi-wheeled passive sliding}
Underactuated aerial vehicles face limitations in passive sliding tasks, particularly when sliding in arbitrary directions (i.e., omnidirectional sliding motion), due to the inherent coupling between linear and angular dynamics. Instead, fully-actuated aerial vehicles enable decoupled linear and angular motion control with 6-\ac{dof} wrench (i.e., forces and torques) generation \citep{fully_review}. The full-actuation of these aerial vehicles is often realized using either fixed tilted propellers \citep{fully2015,fully_nikou, fully_2018,antonio2019, park} or actively tiltable propellers \citep{watson2022, evan2014, fully2017, voliro, karen2019}. Fully-actuated aerial vehicles with three or four non-actuated omniwheels rigidly attached to the vehicle body allow omnidirectional sliding motion without changing the body orientation. This motivates the use of such aerial systems for passive sliding tasks.

\subsection{Challenges}
The most common control methods for tilt-rotor aerial vehicles in physical interactions are impedance control and hybrid motion/force control \citep{anibal2022}. Between them, impedance control is more widely used since switching controllers between free-flight and interactions can be avoided \citep{anibal2022}.
In aerial systems, there are common systematic uncertainties caused by modeling errors in thrust estimation, \ac{com} offsets, and sensor noise. 
The effects of these uncertainties are often neglected for free flight, however, they can be critical for multi-wheeled passive sliding tasks which require full contact among all wheels and the work surface. This underscores the need for a comprehensive investigation of sliding tasks using aerial robots with multiple non-actuated wheels rigidly attached. 
%

\section{Contributions}
\begin{figure}[!t]
\centering
\includegraphics[width=\columnwidth]{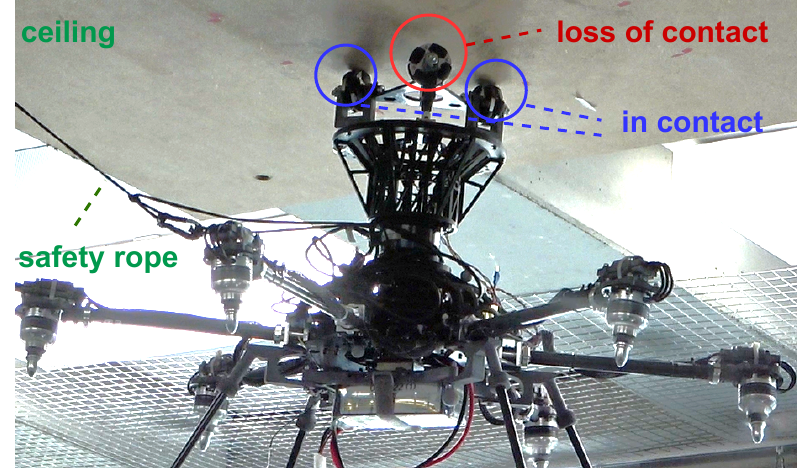}
\caption{Image of a tip-over during a push-and-slide task with a fully-actuated aerial vehicle in a non-actuated three-wheel configuration. For the full action, please see the attached video$^1$. The aerial vehicle is equipped with three omniwheels at the end-effector tip. A wooden board is mounted on the ceiling as the work surface to be slid on. In this tip-over image, two wheels (marked with blue circles) are in contact with the work surface while one (marked with a red circle) loses contact. A safety rope is used to reduce platform damage in risky scenarios.}
\label{fig:tipover_imag}
\end{figure}
To understand the current limitations, we examined the case of a fully-actuated aerial vehicle, with three non-actuated omniwheels rigidly attached to its body using a baseline impedance controller as described in \citep{karen2019}. A critical issue identified during the sliding experiments was the loss of wheel contact, which we define as the \textit{tip-over} problem in our context (see Fig.~\ref{fig:tipover_imag}), even though in mobile manipulation the term \enquote{tip-over} typically refers to a more severe instability where the robot completely overturns \citep{ZMP}. In our case, any situation where the robot loses wheel contact is considered a tip-over.

To address this issue, we conduct a detailed study on applying tip-over stability criteria, originally developed for mobile manipulators, on aerial systems. In addition, we propose two solutions from both mechanical design and control perspectives to improve task performance, utilizing different tip-over stability criteria. Experimental data informs the derivation of mechanical design guidelines and the creation of a simulator that closely captures real-world conditions. From a control standpoint, we introduce a novel pressure-sensing-based control framework that ensures reliable task performance even during sliding maneuvers with small pushing forces and in the presence of systematic uncertainties. The simulator is then used to test and benchmark the effectiveness and robustness of the proposed approaches in high-risk scenarios with measured uncertainties.

With this work, we aim to advance the sliding technology with aerial robots in non-actuated multi-wheel configurations by providing in-depth analysis and innovative solutions.
\section{Paper Outline}
The rest of the paper is organized as follows. Sec.\ref{sec:case_study} presents the push-and-slide experiments conducted using a fully-actuated aerial vehicle equipped with three non-actuated wheels. The experiments reveal the tip-over issue shown in the provided video\footnote{The video is available at \url{https://youtu.be/BfNPLJq_JDs}.} and data\footnote{The corresponding experimental dataset \citep{data} is available at \url{https://data.mendeley.com/datasets/8wzk89df89/1}.}. Sec.\ref{sec:analysis} introduces related work on tip-over stability criteria developed for mobile manipulators and their application on aerial systems. Sec.\ref{sec:guidelines} derives end-effector design guidelines based on the analysis in Sec.\ref{sec:analysis} to mitigate the identified tip-over issue. Sec.\ref{sec:tipover_control} proposes a novel control framework that uses pressure-sensing on each wheel to resolve the tip-over issue even in highly risky scenarios. In Sec.\ref{sec:validation}, the proposed mechanical and control approaches are tested and benchmarked in simulation environments that closely capture the real-world conditions of the experiments in Sec.\ref{sec:case_study} to validate their effectiveness. Finally, Sec.\ref{sec:conclusion} concludes the article with a summary of the findings and suggests possible directions for future research.

\section{Problem Formulation}\label{sec:case_study}
In this section, we introduce the studied fully-actuated aerial platform in Fig.~\ref{fig:tipover_imag} and the tip-over issue identified from the push-and-slide experiments.
\subsection{Platform Description}\label{sec:platform}
\begin{figure}[!t]
\centering
\includegraphics[width=\columnwidth]{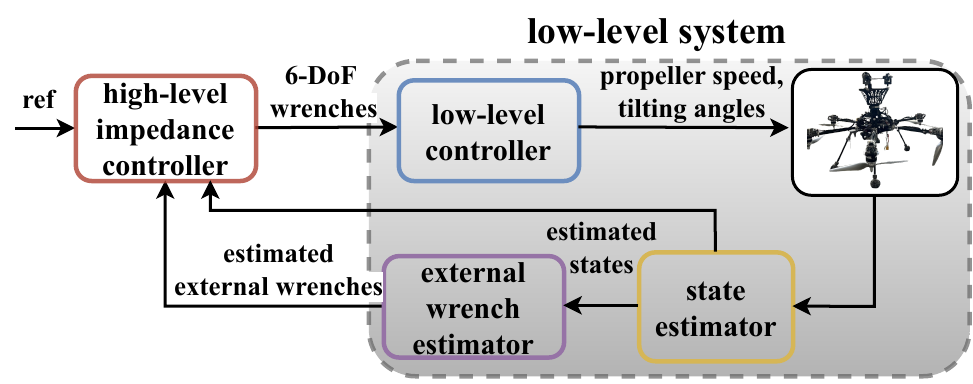}
\caption{Baseline control framework of the fully-actuated aerial vehicle that is composed of a high-level impedance controller and a low-level system.}
\label{fig:baseline}
\end{figure}

\begin{figure}[!t]
\centering
\begin{subfigure}{0.45\columnwidth}
    \centering
    \includegraphics[width=\columnwidth]{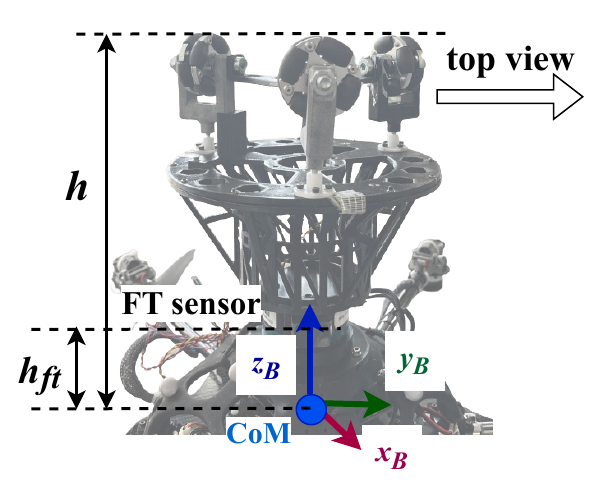}
    \caption{}
    \label{fig:wheel_num}
\end{subfigure}
\begin{subfigure}{0.45\columnwidth}
    \centering
    \includegraphics[width=\columnwidth]{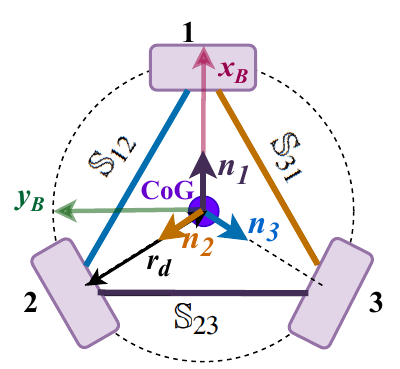}
    \caption{}
    \label{fig:omni}
\end{subfigure}
\caption{\new{The (a): end-effector equipped with three omniwheels, body frame at \ac{com}: $\mathcal{F}_B:=\{O_B;\bm{x}_B,\bm{y}_B,\bm{z}_B\}$. The (b): top view of the wheel layout, $\bm{n_1}$, $\bm{n_2}$, $\bm{n_3} \in \mathbb{R}^2$ are normal vectors of planes $\mathbb{S}_{23}$, $\mathbb{S}_{31}$ and $\mathbb{S}_{12}$ introduced in Sec.\ref{sec:3d}.}}
\label{fig:layout}
\end{figure}

The fully-actuated aerial vehicle is equipped with tiltable propellers. The baseline control framework of this platform uses a high-level 6-\ac{dof} impedance controller and a low-level controller that allocates the 6-\ac{dof} wrenches from the high-level controller into propeller rotating speed and the angles of tiltable propeller arms, see Fig.~\ref{fig:baseline}. \new{We call the system together with the low-level controller, a state estimator, and an external wrench estimator - the \textit{low-level system}. The state estimator fuses data from the Vicon motion capture system with on-board inertial measurement unit (IMU) measurements. The estimated states are used in an external wrench estimator that follows a momentum-based approach introduced in \citep{estimation}.} With the given control framework the platform can exert forces and torques in all directions at any body orientation, in addition to the gravity compensation. With the body frame $\mathcal{F}_B:=\{O_B;\bm{x}_B,\bm{y}_B,\bm{z}_B\}$ attached to its \ac{com} as in Fig.~\ref{fig:wheel_num}, an end-effector is rigidly attached to the aerial vehicle along the axis $\bm{z}_B$ which is defined as the interaction axis. The end-effector is equipped with three omniwheels and the wheels are equally distributed with the same \new{wheel distance} $r_d=$\SI{0.084}{\meter} from the \ac{cog} of the end-effector tip, see Fig~\ref{fig:omni}. Moreover, they are placed with a 120$\degree$ angle between each pair to allow sliding motion in an arbitrary direction along a flat surface.
Let $h=$\SI{0.3}{\meter} be the distance from \ac{com} to end-effector tip along $\bm{z}_B$. A 6-\ac{dof} FT (force and torque) sensor is mounted between the end-effector and the aerial vehicle. The sensor frame orientation coincides with the body frame orientation with an offset $h_{ft}=$\SI{0.12}{\meter} in axis $\bm{z}_B$ between the mounting point of the force sensor and $O_B$, see Fig.~\ref{fig:wheel_num}. The FT sensor is used to measure low-level system uncertainties and is not used in the control scheme. 
\subsection{Tip-over issue}\label{sec:issue}
The experimental setup is illustrated in Fig.\ref{fig:tipover_imag}, and the aerial vehicle is commanded to execute push-and-slide tasks on the wooden board with the baseline controller in Fig.~\ref{fig:baseline}. The orientation of the wooden board is unknown. In the case of multi-wheeled passive sliding on an unknown surface, using a fully-actuated aerial vehicle with 6-\ac{dof} impedance control suffers from systematic uncertainties and inaccurate attitude references. These issues result in attitude errors leading to initial misalignment between the end-effector tip and the work surface, as depicted at the beginning of the video\footnotemark[1]. Additionally, wheel contact during sliding is not always ensured, both of which contribute to the robot tip-over as in Fig.\ref{fig:tipover_imag} and highlighted throughout the video\footnotemark[1]. While these issues are demonstrated with a specific fully-actuated platform, they are indicative of broader challenges faced by generic aerial systems when executing push-and-slide tasks with the existing control framework. In the experiment, initial tip-over upon contact was addressed by manually adjusting the aerial vehicle's orientation: a process that is time-consuming and prone to instability due to potential human error. Tip-over during sliding frequently results in low-quality measurements in NDT or reduced precision in other applications. These challenges underscore the need for further investigation into the theoretical conditions for tip-over stability in multi-wheeled sliding operations involving aerial vehicles.

\section{Tip-Over Stability Study of Aerial Systems}\label{sec:analysis}
This section presents various tip-over stability criteria developed for mobile manipulation and their application on aerial systems. Based on the experimental data, we selected and applied a criterion associated with the aerial system examined in Sec.~\ref{sec:case_study}. Finally, we discuss mitigation strategies related to tip-over based on the presented stability criteria.
\subsection{Existing Tip-over Stability Criteria}\label{sec:tipover_stability}
The tip-over problem has been broadly studied for over twenty years on mobile robots and several tip-over stability criteria are developed. \citet{ZMP} and \citet{ZMP_control} presented the \ac{zmp} criterion-based tip-over stability measure for mobile manipulators, which cannot tackle the change of \ac{com}. \citet{energy} reported the energy-based stability criterion which has the risk of real-time instability due to its complex computation~\citep{ding2019}. \citet{moment_height} introduced the moment-height tip-over stability measure which requires the computation of the moment of inertia concerning each side of the support polygon. \citet{force_angle} \citep{force_angle_book} presented the force-angle tip-over stability measure which captures the change of \ac{com} in contrast to the \ac{zmp} criterion and is easy to compute. 

Among the aforementioned tip-over stability criteria, the force-angle stability measure requires relatively low computation cost and is effective in tip-over prediction for mobile manipulators. However, considering aerial systems which are often subjected to more systematic uncertainties and higher risk of real-time instability compared to mobile robots, the effectiveness of the force-angle stability measure on fully-actuated aerial systems needs further investigation. \citet{support_force} introduced a normal-force-based tip-over stability criterion which can explicitly measure the tip-over stability using force sensors. The criterion is then used for online tip-over prevention by limiting the system motion. And yet, in wheeled mobile manipulator studies, there is a lack of active control approaches that help the system recover from tip-overs and actively generate control wrenches to avoid tip-overs even during sliding maneuvers. This is because the wheeled mobile manipulators often do not have full actuation in 6-\ac{dof}. Fully-actuated aerial systems, however, offer such capabilities. Moreover, the normal forces can be measured with only pressure sensors attached to wheels at a low cost, and the associated stability criterion is straightforward to implement which makes it a good candidate for tip-over control.
\begin{figure*}[t]
   \centering
   \begin{subfigure}{0.23\textwidth}
   \centering
\includegraphics[trim={0.0cm 0.0cm 0.0cm 0.0cm},clip,width=\columnwidth]{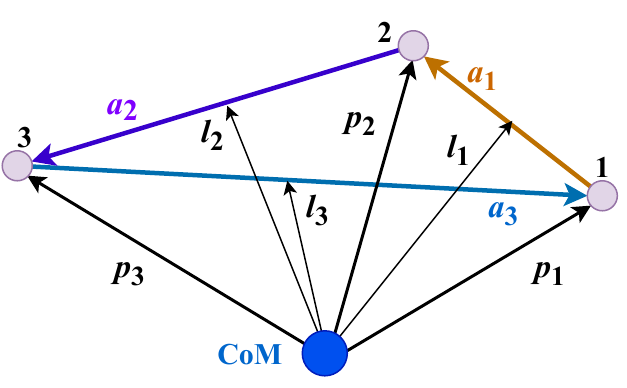}
    \caption{}
    \label{fig:vector1}
    \end{subfigure} \hfill
    \begin{subfigure}{0.25\textwidth}
    \centering
    \includegraphics[trim={0.0cm 0.0cm 0.0cm 0.0cm},clip,width=\columnwidth]{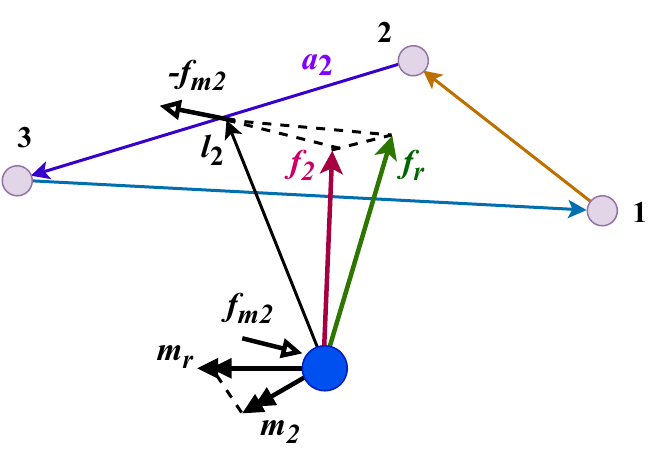}
    \caption{}
    \label{fig:vector2}
\end{subfigure}
     \begin{subfigure}{0.25\textwidth}\centering
    \includegraphics[trim={0.0cm 0.0cm 0.0cm 0.0cm},clip,width=\columnwidth]{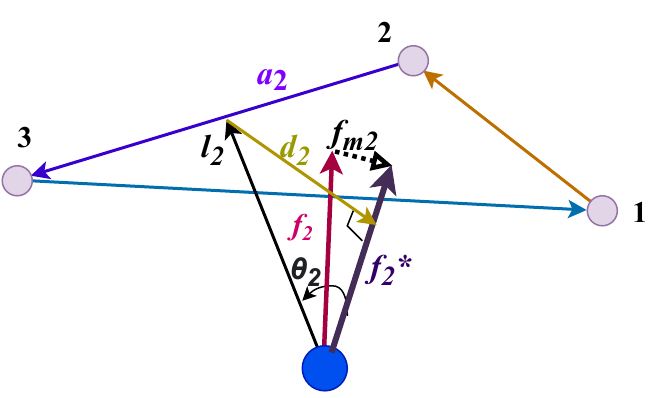}
    \caption{}
    \label{fig:vector3}
\end{subfigure}
 \begin{subfigure}{0.25\textwidth}\centering
    \includegraphics[trim={0.0cm 0.0cm 0.0cm 0.0cm},clip,width=\columnwidth]{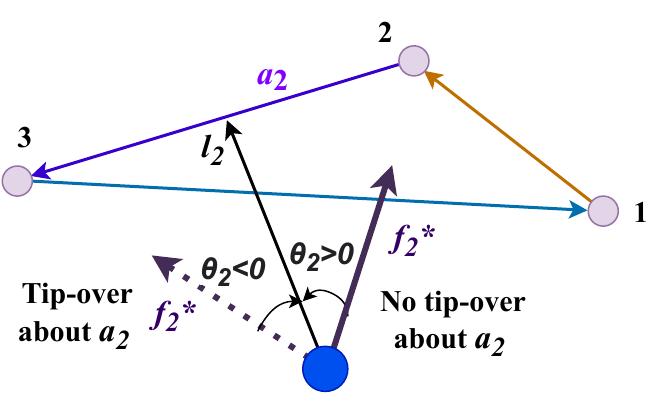}
    \caption{}
    \label{fig:vector4}
\end{subfigure}
\caption{Force-angle measure definition, (a): support pattern with tip-over axes and their normal vectors; (b): force and torque vectors that participate in tip-over instability; (c): the resultant force vector and the candidate angle; (d) the candidate angle value and corresponding tip-over status.}\label{fig:force_angle}
\end{figure*}

\subsection{Force-angle tip-over stability measure}\label{sec:force_angle}
This section presents the offline force-angle stability measure from physical experiments with the fully-actuated aerial vehicle in Sec.~\ref{sec:platform} following the methods in \citep{force_angle, force_angle_book}.
During the physical experiments, the platform came into contact with a flat surface on the ceiling by pushing toward it and then slid on the surface. While sliding, tip-over instability arises when the robot body undergoes rotation, leading to a decrease in the number of contact points. Eventually, all remaining contact points align along a single line, known as the tip-over axis. For the studied platform with three omniwheels in Fig.~\ref{fig:wheel_num}, tip-over axes are defined as the lines that join all contact points of the support polygon. The force-angle stability measure monitors the tip-over stability using a resultant force vector and a candidate angle about each tip-over axis. This force vector captures the effects of forces and torques acting on the \ac{com} of the system that contribute to the tip-over instability about the corresponding tip-over axis. The candidate angle indicates the tip-over status about the corresponding tip-over axis under the effects of the resultant force vector. The tip-over stability measure of the whole system is thus given by a measure of the most crucial tip-over axis around which the tip-over occurs first.

In the following, we introduce a support pattern associated with the support polygon and its tip-over axes. The support pattern is considered as a stable region within which no tip-over occurs. Furthermore, the resultant force vector definition is presented for the studied system and a candidate angle that presents the tip-over status is defined based on the location of the resultant force vector \ac{wrt} the support pattern. The resulting data from experiments are used to display the force-angle stability measure of the whole system and derive tip-over avoidance guidelines for hardware design. All vectors in this section are defined in the body frame $\mathcal{F}_B$.

\subsubsection{The Support Pattern}
For a consistent formulation despite the robot body orientation, we define that the wheel number $i$ increases from $1$ to $3$ following the right-hand rule with the thumb aligning with $\bm{z}_B$ as in Fig.~\ref{fig:vector1}. The contact points from three wheels form a support polygon on the work surface in the shape of a triangle which introduces three tip-over axes $\bm{a}_i \in \mathbb{R}^3, \ i=1,2,3$. Let $\bm{p}_i$ be the position vector of the contact point on wheel $i$ \ac{wrt} $\mathcal{F}_B$. We define that for $i=3$, $\bm{p}_{i+1}:=\bm{p}_1$. Therefore, for $i=1,2$, $\bm{a}_i=\bm{p}_{i+1}-\bm{p}_i$, and $\bm{a}_3=\bm{p}_1-\bm{p}_3$. We denote~~$\widehat{}$~~as the normalization operator where $\widehat{\bm{a}}=\displaystyle{\bm{a}}/{||\bm{a}||}$ is the unit vector of an arbitrary vector $\bm{a}\neq \bm{0}$. Let $\bm{l}_i$ be the normal vector of $\bm{a}_i$ which passes through the \ac{com} and is given by:
$\bm{l}_i=(\mathbb{I}_{3\times 3}-\widehat{\bm{a}_i}\widehat{\bm{a}_i}^{\top})\bm{p}_{i+1}$, where $\mathbb{I}_{3\times 3}$ is the identity matrix. The support pattern of the studied fully-actuated aerial vehicle is \new{constructed} by the defined normal vectors $\bm{l}_i, i=1,2,3$, and $\bm{l}_i$ is later used to derive the candidate angle to assess the tip-over status about tip-over axis $\bm{a}_i$.
\subsubsection{The Resultant Force Vector}
To find the resultant force vector of the system about each \new{tip-over axis} $\bm{a}_i$, we first define the forces and torques acting on the system \ac{com}, which participate in tip-over instability via the system's equations of motion. The equations of motion of the system expressed in the body frame are written as:
\begin{equation}\label{eq:em}
\bm{M}\dot{\bm{v}}+\bm{C}\bm{v}+\begin{bmatrix}
    \bm{g}^B\\\bm{0}_3
\end{bmatrix}=\bm{w}_a+\bm{w}_e,
\end{equation}
where $\bm{v} \in \mathbb{R}^{6}$ is the stacked linear and angular velocity, $\bm{M}$ and $\bm{C} \in \mathbb{R}^{6 \times 6}$ are the mass and Coriolis matrices, $\bm{g}^B,\ \bm{f}^\top_a,\ \bm{\tau}^\top_a \in \mathbb{R}^{3}$ are the gravity force vector, actuation force vector, and actuation torque vector respectively. $\bm{w}_a=\begin{bmatrix}
    \bm{f}^\top_a& \bm{\tau}^\top_a
\end{bmatrix}^{\top}$ and $\bm{w}_e \in \mathbb{R}^6$ are the actuation wrenches and external wrenches. We assume that only the contact forces between the end-effector and the work surface contribute to the external wrenches. 
Being different from mobile manipulators that usually have actuated wheels, for passive sliding with aerial vehicles, the aerial system generates actuation forces to enable the sliding motion and push towards the work surface. Therefore, the forces and torques acting on the aerial system \ac{com} that participate in tip-over instability according to \citep{force_angle} are defined as:
\begin{equation}\label{eq:fr_mr}
    \bm{f}_r =\bm{f}_a-\bm{g}^B,\ \bm{m}_r = \bm{\tau}_a.
\end{equation}
For a given tip-over axis $\bm{a}_i$, only the components of $\bm{f}_r$ and $\bm{m}_r$ that act about $\bm{a}_i$ would contribute to tip-over instability about $\bm{a}_i$. As shown in Fig.~\ref{fig:vector2}, the components of $\bm{f}_r$ and $\bm{m}_r$ about $\bm{a}_i$ are given by:
$
    \bm{f}_i=(\mathbb{I}_{3\times3}-\widehat{\bm{a}}_i\widehat{\bm{a}}_i^{\top})\bm{f}_r$ and $\ \bm{m}_i =(\widehat{\bm{a}}_i\widehat{\bm{a}}_i^{\top})\bm{m}_r$.
Furthermore, the effects of $\bm{m}_i$ on tip-over instability about $\bm{a}_i$ need to be represented in the form of equivalent forces. To do so, we introduce an equivalent force couple, i.e., two force vectors acting on different locations, to replace the torque vector $\bm{m}_i$. With infinite possible force locations and directions, a reasonable choice introduced by \citep{force_angle} is to use a force couple $(-\bm{f}_{mi},\bm{f}_{mi})$, where $-\bm{f}_{mi}$ passes through the cross point of $\bm{a}_i$ and $\bm{l}_i$, and $\bm{f}_{mi}$ passes through $O_B$, see Fig.~\ref{fig:vector2}.
The member of the force couple acting at the \ac{com} is given by: $\bm{f}_{mi}=\frac{\widehat{\bm{l}}_i \times \bm{m}_i}{||\bm{l}_i||}$. Finally, as shown in Fig.~\ref{fig:vector3}, the resultant force vector about tip-over axis $\bm{a}_i$ that captures both effects of forces and torques is given by:
\begin{equation}
    \bm{f}_i^*=\bm{f}_i+\bm{f}_{mi}.
\end{equation}
\subsubsection{Stability Measure of the Fully-Actuated Aerial Vehicle} \label{sec:measure}
With the defined normal vector $\bm{l}_i$ and the resultant force vector $\bm{f}_i^*$ about each axis $\bm{a}_i$ and their unit vectors $\widehat{\bm{l}}_i, \widehat{\bm{f}}_i{}^*$, the previously mentioned candidate angle used for stability measure is defined as $\theta_i=\sigma_i cos^{-1}(\widehat{\bm{f}}_i{}^*,\widehat{\bm{l}}_i) \in [-\pi,\pi]$, see Fig.~\ref{fig:vector4}. $\sigma_i$ is the sign of $\theta_i$, which signals if the force vector $\bm{f}_i^*$ is inside the support pattern, and is given by:
\begin{equation}
    \sigma_i=\begin{cases}
    +1 & (\widehat{\bm{f}}_i{}^* \times \widehat{\bm{l}}_i)\cdot \widehat{\bm{a}}_i>0,\\ -1 & \text{otherwise}.
\end{cases}
\end{equation}
When the angle has a positive sign, it indicates that the force vector $\bm{f}_i^*$ is inside the support pattern, and tip-over about axis $\bm{a}_i$ is in progress if the sign is negative. The critical case is $\theta_i=0$ when $\bm{f}_i^*$ coincides with $\bm{l}_i$. Among all the angles $\theta_i$ for $i=1,2,3$, the smallest $\theta_i$ indicates the most crucial tip-over axis around which the tip-over occurs first. The force-angle stability measure of the whole system is therefore given by:
\new{\begin{equation}\label{eq:alpha}
    \alpha=\min_{i\in1,2,3}(\theta_i \cdot ||\bm{d}_i|| \cdot ||\bm{f}_i^*||),
\end{equation}}
with $\bm{d}_i=-\bm{l}_i+(\bm{l}_i^{\top} \cdot \widehat{\bm{f}}_i{}^*)\widehat{\bm{f}}_i{}^*$ being the minimum length vector from the tip-over axis $\bm{a}_i$ to $\bm{f}_i^*$ (for details, please refer to \citep{force_angle_book}) as in Fig.~\ref{fig:vector3}.
\begin{figure}[!t]
\centering
\includegraphics[width=\columnwidth]{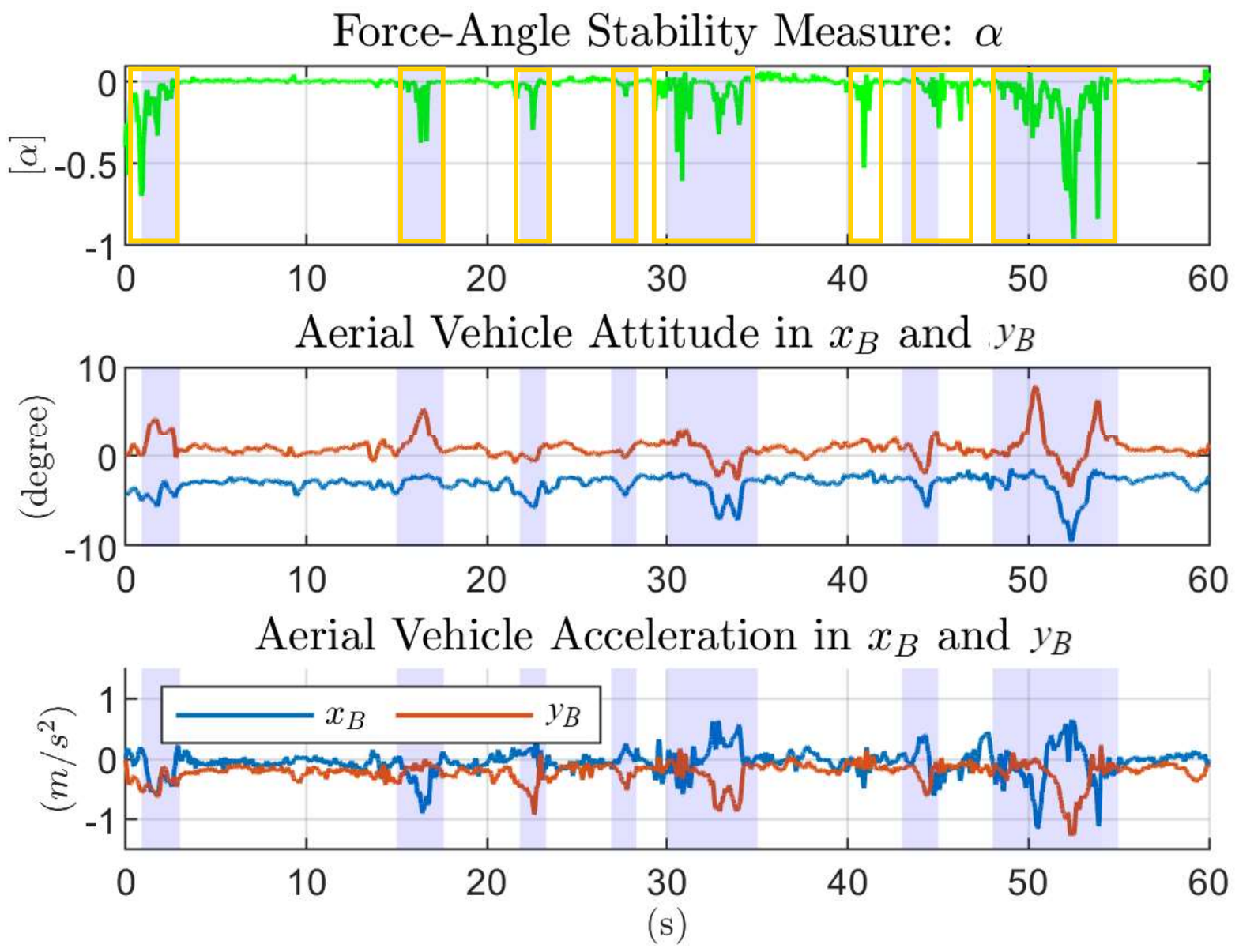}
\caption{From top to bottom: force-angle stability measure, aerial vehicle attitude along $\bm{x}_B$ and $\bm{y}_B$, aerial vehicle acceleration along $\bm{x}_B$ and $\bm{y}_B$; shaded area: visually captured tip-over incidents from the experiment associated with changes of the aerial vehicle attitude in $\bm{x}_B$ and $\bm{y}_B$; yellow rectangles: expected tip-overs based on the force-angle stability measure.}
\label{fig:measure}
\end{figure}

Control wrenches $\bm{w}_a$ in \eqref{eq:em} generated by the propulsion system of the platform, system linear acceleration, and aerial vehicle attitude can be obtained from the experimental data\footnotemark[2]. With \eqref{eq:fr_mr} and \eqref{eq:alpha}, the force-angle stability measure for the fully-actuated aerial vehicle during passive sliding is displayed in Fig.~\ref{fig:measure} along with the aerial vehicle attitude and linear acceleration in $\bm{x}_B$ and $\bm{y}_B$.
The attitude and linear motion in $\bm{z}_B$ do not contribute to tip-over.
The yellow rectangles in the top plot of Fig.~\ref{fig:measure} indicate expected tip-overs based on the force-angle stability measure criterion, while the shaded area presents visually captured tip-over incidents from the experiment. During the experiment, tip-overs were associated with evident changes in aerial vehicle orientation along $\bm{x}_B$ and $\bm{y}_B$, as shown in the middle plot of Fig.~\ref{fig:measure}. 
\subsection{Discussion}
The force-angle stability measure effectively captures most of the tip-over incidents during the experiments, proving its reliability in assessing potential instances of robot tip-over instability. The uncertainties from the low-level system can affect the accuracy of the tip-over prediction, which makes the force-angle stability measure less robust to be used in real-time tip-over control, but an efficient tool to evaluate the likelihood of tip-over instability with the system setup.

Tip-overs generally occurred when there was an explicit increase of the sliding acceleration (i.e., the linear acceleration along $\bm{x}_B$ and $\bm{y}_B$) magnitude. In regions where no tip-over is detected \new{by the force-angle stability criterion}, the measured stability has very small values, being slightly above the critical case when $\alpha=0$. Being too close to the critical case heightens the system's sensitivity to the effects of systematic uncertainties or disturbances. As a result, the likelihood of tip-over instability increases when subjected to these effects. Hence, to prevent tip-overs in the presence of uncertainties and potential disturbances, it is essential to increase the system's stability measure effectively. This entails establishing a substantial margin between the stability measure value and the critical value.

\section{Mechanical Design Guidelines}
\label{sec:guidelines}
In this section, we present hardware design guidelines that aid in increasing the tip-over stability measure and avoiding the critical case. Based on the definition in Sec.~\ref{sec:measure}, the key to avoiding tip-overs is to keep the force vector $\bm{f}_i^*$ inside the support pattern. Increasing the force-angle stability measure to gain a bigger stability margin can be achieved by either constraining the resultant force vector or enlarging the support pattern.
During passive sliding, the fully-actuated aerial vehicle has to exert a pushing force along the interaction axis $\bm{z}_B$ while supplying also the lateral forces along $\bm{x}_B$ and $\bm{y}_B$ to enable sliding motion. As the sliding acceleration magnitude rises, the lateral forces also increase, resulting in a simultaneous decrease in the stability measure as shown in Fig.~\ref{fig:measure}. Hence, either reducing the sliding acceleration or increasing the pushing force magnitude can improve the tip-over stability without changing the support pattern. Nevertheless, this approach imposes constraints on the application range where varied sliding accelerations are demanded and entails increased energy consumption when generating higher pushing forces. Consequently, we rather focus on hardware guidelines to enlarge the support pattern. This can be achieved by either increasing the support polygon (i.e., rising $r_d$) or reducing the distance from the \ac{com} to the end-effector tip (i.e., decreasing $h$).
With the same acceleration and force data used in Fig.~\ref{fig:measure}, we increase $r_d$ to $2r_d$, $5r_d$, and reduce $h$ to $h/2$, $h/5$. The effectiveness of increasing the stability measure by enlarging the support pattern is shown in Fig.~\ref{fig:a_up}.
\subsection{Limitations}
\begin{figure}[!t]
\centering
\includegraphics[width=\columnwidth]{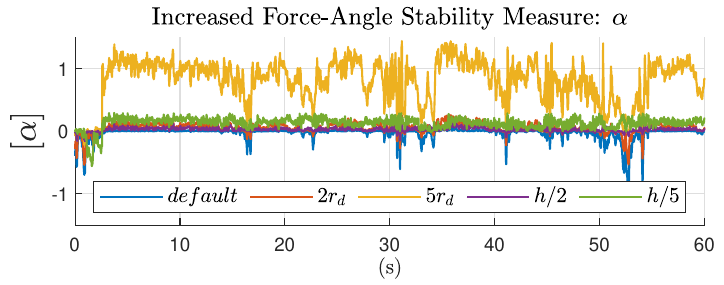}
\caption{Increased stability measure by increasing $r_d$ to $2r_d$, $5r_d$ with constant $h$, and reducing $h$ to $h/2$, $h/5$ with constant $r_d$.}
\label{fig:a_up}
\end{figure}

The guidelines based on the force-angle stability measure can help us prevent tip-overs during sliding via hardware design and avoid investigation in control design.
However, scaling up $r_d$ or scaling down $h$ twice does not significantly improve the stability measure as shown in Fig.~\ref{fig:a_up}. In practical terms, doubling the end-effector tip size results in a bulky design relative to the aerial vehicle size and mass. Additionally, there is a lower limit for $h$ to safeguard the propellers from contacting the work surface. Even with $h/5$, the stability measure still approaches the critical value $0$ for riskier scenarios when sliding maneuvers occur.
Considering the platform size in practice, the end-effector design with a wheel distance of $2r_d$ is evaluated in Sec.~\ref{sec:validation}. Moreover, misalignment between the end-effector tip and the work surface at initial contact remains unresolved with the baseline impedance control. These limitations motivate the need for a more robust solution to enhance the baseline control framework for passive sliding tasks.

\section{Control Framework to Enhance Passive Sliding} \label{sec:tipover_control}
\begin{figure}[!t]
\centering
\includegraphics[width=\columnwidth]{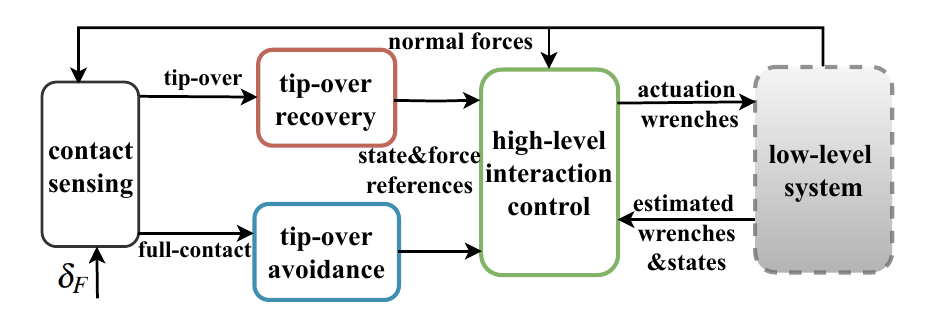}
\caption{Control framework for enhancing passive sliding with the fully-actuated aerial vehicle. Contact sensing identifies the system's tip-over status. Tip-over recovery mode addresses the initial misalignment and tip-over avoidance mode preserves the wheel contact while sliding. The high-level interaction controller uses normal forces at each wheel to preserve wheel contact while keeping the system position. The actuation wrenches output from the high-level controller are fed to the low-level system described in Fig.~\ref{fig:baseline} of Sec.~\ref{sec:platform}.}
\label{fig:control}
\end{figure}

To enhance sliding performance, we introduce a novel control framework, as depicted in Fig.\ref{fig:control}. This framework consists of a contact sensing process, a low-level system shown in Fig.\ref{fig:baseline}, and a high-level interaction controller. The contact sensing process monitors the system's tip-over status, triggering two modes of action: tip-over recovery mode, which addresses initial misalignment, and tip-over avoidance mode, which maintains wheel contact during sliding. The high-level interaction controller employs a normal-force-based tip-over stability criterion \citep{support_force} that precisely measures tip-over instability despite systematic uncertainties (as discussed in Sec.~\ref{sec:tipover_stability}). This framework requires the use of pressure sensors to measure the normal forces acting on the wheels.

To simplify the problem, we first examine a two-wheeled planar system and introduce the high-level interaction control design for this system. We then present the method for extending this control design from the planar system to a three-wheeled system in 3D space, as shown in Fig.~\ref{fig:tipover_imag}.
\begin{figure}[!t]
\centering
\includegraphics[width=\columnwidth]{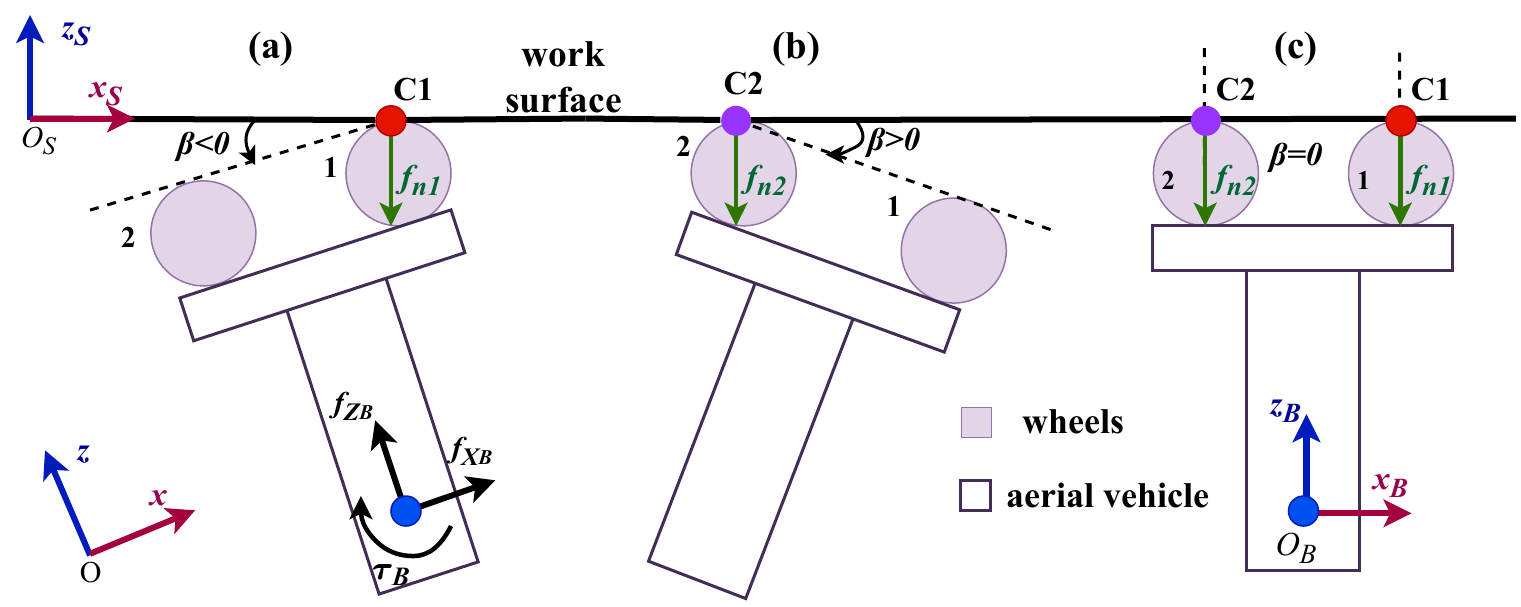}
\caption{Planar two-wheeled system and coordinate frames, (a): tip-over with $\beta<0$, wheel 1 in contact; (b): tip-over with $\beta>0$, wheel 2 in contact; (c): full-contact with $\beta=0$.}
\label{fig:force}
\end{figure}
\begin{table}[!t]
\caption{Notation}
\begin{tabularx}{\columnwidth}{p{0.08\textwidth}X}
\toprule
$i$ & wheel number\\
 $C_i$ & the contact point on wheel $i$\\
 $m_B$ & the total mass of the system\\
\multicolumn{2}{l}{\underline{Coordinate frames:}}\\ 
  $\mathcal{F}'_w$& the world frame of any motion, $\mathcal{F}'_w:=\{O;\bm{x},\bm{z}\}$\\ 
  $\mathcal{F}'_S$& the work frame attached to the work surface, $\mathcal{F}'_S:=\{O_S;\bm{x}_S,\bm{z}_S\}$\\
  $\mathcal{F}'_B$& the body frame attached to the \ac{com} of the system, $\mathcal{F}'_B:=\{O_B;\bm{x}_B,\bm{z}_B\}$\\
  \multicolumn{2}{l}{\underline{State variables:}}\\ 
  $x$& the position of \ac{com} w.r.t. $O$ along $\bm{x}_B$\\
 $\beta$ & the orientation of the body frame w.r.t. the work frame being positive clockwise, $\beta <0$ when only wheel 1 is in contact, $\beta >0$ when only wheel 2 is in contact, and $\beta=0$ for full-contact, see Fig.~\ref{fig:force}\\ 
\multicolumn{2}{l}{\underline{Contact forces:}}\\ 
     $f_{ni}$ & the normal force acting on wheel $i$, $f_{ni}\geq0$\\
     \multicolumn{2}{l}{\underline{Actuation wrenches:}}\\ 
     $f_{X_B}$& actuation force acting on \ac{com} along $\bm{x}_B$, $f_{X_B} \in \mathbb{R}$\\
     $f_{Z_B}$& actuation force acting on \ac{com} along $\bm{z}_B$, $f_{Z_B} \in \mathbb{R}$\\
    $\bm{f}_B$& the actuation linear force vector expressed in body frame, $\bm{f}_B=\begin{bmatrix}
    f_{X_B}&f_{Z_B}
\end{bmatrix}^{\top} \in \mathbb{R}^2$\\
$\tau_B$& the actuation rotational torque along the axis perpendicular to the 2D plane, being positive clockwise,$\tau_B \in \mathbb{R}$\\
 \bottomrule
 \end{tabularx}\label{table:notation}
\end{table}
\begin{table}[!t]
\caption{Contact Conditions}\label{table:contact}
\begin{center}      
    \begin{tabularx}{\columnwidth}{p{0.16\columnwidth}|p{0.1\columnwidth}|p{0.1\columnwidth}|p{0.18\columnwidth}|X}
     \hline
    Status& $f_{n1}$&$f_{n2}$ &discretized $\beta$ &Control Objective \\ 
     \hline
     \textbf{free flight}  & $<\delta_F$&$<\delta_F$&$-2$ & full-pose control\\
     \hline
     \textbf{tip-over}  & $\geq\delta_F$&$<\delta_F$&$-1$ & tip-over recovery\\ \hline
     \textbf{tip-over}  & $<\delta_F$&$\geq\delta_F$&$+1$ & tip-over recovery\\ \hline
     \textbf{full-contact}  & $\geq\delta_F$&$\geq\delta_F$&$0$& tip-over avoidance\\ \hline
\end{tabularx}
\end{center}
\end{table}
\subsection{Planar Two-Wheeled System}
 To capture the main characteristics of passive sliding with the studied fully-actuated aerial vehicle on a flat work surface, we present a simplified two-wheeled system model in a 2D plane. The force and torque generation ability of the platform in Sec.\ref{sec:platform} allows us to neglect gravity force in the simplified system, and with the focus on high-level control design, we assume that the system can directly supply actuation forces in any direction of the 2D plane and the actuation torque along the axis perpendicular to the plane. 
Now consider a two-wheeled system as in Fig.~\ref{fig:force}, in which wheel 1 and wheel 2 have their centers rigidly connected to an aerial vehicle which is assumed to be a rigid body.
Two wheels can freely rotate around the axis perpendicular to the 2D plane. When both wheels are in contact with the work surface, we define such contact condition as \textit{full-contact}. On the other hand, when there is only one wheel in contact, it is referred to as a tip-over.
The notations used in the following sections are displayed in Table~\ref{table:notation}. Without including uncertainties and other external disturbances apart from the contact wrenches in the system modeling, the system's equations of motion expressed in the body frame are given by:
\begin{equation}\label{eq:2wheelmodel}
    \bm{M}_S\dot{\bm{v}}_S+\bm{C}_S\bm{v}_S=\bm{w}_B+\bm{w}_{C},
\end{equation}
where $\bm{v}_S \in \mathbb{R}^{3}$ is the stacked linear and angular velocity, $\bm{M}_S$ and $\bm{C}_S \in \mathbb{R}^{3 \times 3}$ are the  mass and Coriolis matrices, $\bm{w}_B=\begin{bmatrix}
    \bm{f}_B^\top& \tau_B
\end{bmatrix}^{\top}$ and $\bm{w}_{C} \in \mathbb{R}^3$ are the actuation wrenches and contact wrenches respectively. 
As shown in Fig.~\ref{fig:force}, a tip-over with $\beta<0$ indicates when only wheel 1 is in contact, a tip-over with $\beta>0$ indicates when only wheel 2 is in contact, and $\beta=0$ indicates when the system is in full-contact. During tip-overs, the system is only subjected to the contact forces acting on one contact point $C_i$ depending on which wheel is in contact. In full-contact, the contact forces act on both wheels.
\subsection{Contact Sensing}\label{sec:tip-over}
In this section, we present the contact sensing process of Fig.~\ref{fig:control} to identify different contact conditions using the normal-force-based tip-over stability criterion \citep{support_force} for the modeled two-wheeled system. The normal forces acting on wheels are used to identify the contact conditions between the system and the work surface. When the normal force on a wheel goes to zero, it means this wheel is detached from the work surface. To increase the robustness of tip-over detection, instead of zero, we use a positive margin $\delta_F$. With $\delta_F$, we define four contact conditions, as shown in Table ~\ref{table:contact} where $\beta$ is discretized. We set $\beta=-2$ for free flight, $\beta=0$ for full-contact, $\beta=-1$ when wheel 1 is in contact during the tip-over, and $\beta=+1$ when wheel 2 is in contact during the tip-over, such that the corresponding contact conditions are digitalized and can be used in the control design.
This process is then realized in the control framework using a pressure sensor below each wheel that provides contact sensing. Only if full-contact or a tip-over condition is detected, the interaction controller is enabled. When the system is resting in full-contact without sliding by only pushing towards the surface along the interaction axis (i.e., the static contact), ideally the normal forces acting on two wheels are equal without considering uncertainties and other potential disturbances due to symmetric properties \citep{support_force}. When the system tends to tip over, the normal force on one wheel starts to reduce while the difference between the normal forces on two wheels increases. Therefore, we define the static contact case as the nominal contact, where:
\begin{equation}\label{eq:equal}
    f_{n1}=f_{n2}, \ \text{with $f_{n1},f_{n2}\geq\delta_F$}.
\end{equation}
The tip-over stability criterion is used to monitor the contact conditions. We use the nominal contact case to derive the control law in the following section. \new{The \eqref{eq:equal} is however only a sufficient condition of full-contact while having both $f_{n1},f_{n2}\geq\delta_F$ is considered as a necessary condition of full-contact.}
\subsection{High-Level Interaction Control Design}
\label{sec:control}
In this section, we detail the high-level interaction control design in Fig.~\ref{fig:control}. The controller is formulated in body frame $\mathcal{F}'_B$ orientation, as uncertainties about surface orientation render a world frame formulation impractical.
\subsubsection{Action Modes based on Contact Sensing}
When the system is in a tip-over, sending sliding motion references (i.e., linear acceleration, velocity, and position references along $\bm{x}_B$) that are not in parallel with the flat surface often leads to instability. Therefore, when a tip-over is in progress, we set the sliding motion references to zeros until the system gains full-contact. Based on the identified contact conditions in Table~\ref{table:contact}, we introduce two action modes by monitoring the tip-over status: the tip-over recovery and tip-over avoidance modes as in Fig.~\ref{fig:control}. When the system is in a tip-over, the tip-over recovery mode is triggered to recover the system from the tip-over. This mode can address the initial misalignment issue that cannot be handled by the mechanical guidelines provided in Sec.~\ref{sec:guidelines}. When the system is in full-contact, the tip-over avoidance mode preserves the wheel contact. The designed controller provides high-level control outputs described as the actuation wrench $\bm{w}_B$ in the body frame, fed to the low-level system for actuation allocation described in Sec.~\ref{sec:platform}.
\subsubsection{Hybrid Control Design}
With the defined control objectives, we propose a hybrid motion/force control approach for the high-level interaction control design, which we call the \textit{normal-force control}. In the normal-force control scheme, the pushing force along axis $\bm{z}_B$ and the linear motion along axis $\bm{x}_B$ are feedback controlled with force estimation and state estimation from the low-level system. The rotational torque of the 2D plane (being positive clockwise) is controlled using the normal forces measured on both wheels. The direct force control along $\bm{z}_B$ allows the system to exert a specific force on the work surface, and also to ensure physical contact between the system and the work surface. The linear motion control along $\bm{x}_B$ tracks the sliding motion trajectory according to task-specific requirements.
Finally, the torque control handles the tip-over problem. We denote $x^d,\dot{x}^d, \ddot{x}^d$ as sliding motion references along $\bm{x}_B$, and $f^d$ as the force reference along $\bm{z}_B$. We define the linear motion and force tracking errors expressed in the body frame as $e_p=x-x^d$, $e_{d}=\dot{x}-\dot{x}^d$, and $e_{f}=f_{Z_B}^{est}-f^d$, where $f_{Z_B}^{est}$ is the estimated force along $\bm{z}_B$ from the external wrench estimator.

\new{Considering the tip-over stability criterion described in the last section, and aiming at achieving the full-contact described in Table.~\ref{table:contact}, we consider the nominal contact case in \eqref{eq:equal} as the reference condition of the control design. We define the force difference error between the normal force measurements on two wheels as:
\begin{equation}\label{eq:force_error}
    e_n=f_{n1}-f_{n2}.
\end{equation}
Since $e_n=0$ is not necessary for having full-contact, a range of force difference error $e_n$ can be acceptable while achieving full-contact. This property naturally leads to a proportional control design. With acceptable force difference errors, the system saves energy while preserving the full-contact between the end-effector tip and the work surface during interactions. Such control design also compensates for common modeling errors in aerial systems for physical interaction tasks. With a proportional controller, the nominal contact condition thus serves as a guiding condition to let the system rotate in the desired direction towards full-contact.}
We denote $\bm{w}_n=\begin{bmatrix}
    f_x &f_z& \tau_1
\end{bmatrix}^{\top} \in \mathbb{R}^3$ as the control wrenches along $\bm{x}_B$, $\bm{y}_B$ and the rotation axis. With the above definitions, the control wrenches are given by:
 \begin{subequations}
 \begin{equation}\label{eq:motion}
    f_x=m_B\ddot{x}^d-k_pe_{p}-k_de_{d},
    \end{equation}
    \begin{equation}\label{eq:force}
        f_z=f^d-k_fe_f-k_I\int e_f,
    \end{equation}
   \begin{equation}\label{eq:torque}
       \tau_1=k_n e_n,
   \end{equation}
\end{subequations}
where $k_p,k_d,k_f,k_I,k_n$ are positive gains. The final actuation wrenches $\bm{w}_B$ is defined as: $\bm{w}_B=\bm{w}_n+\bm{C}_S\bm{v}_S$. With \eqref{eq:2wheelmodel}, the closed loop dynamics of the system yield:
\begin{equation}
    \bm{M}_S\dot{\bm{v}}_S=\bm{w}_n+\bm{w}_C.
\end{equation}

\subsection{Three-Wheeled System}\label{sec:3d}
In the previous section, we presented the normal-force control design for a simplified planar two-wheeled system, to illustrate how we solve the tip-over issue. In this section, we extend the method from a two-wheeled planar system to the full three-wheeled system shown in Fig.~\ref{fig:wheel_num}, with a focus on torque control.
%
\begin{table}[!t]
\caption{Plane Definition}\label{table:planes}   
\centering
    \begin{tabularx}{\columnwidth}{p{0.15\textwidth}|p{0.08\textwidth}|p{0.08\textwidth}|X} 
     \hline
     Plane& $\mathbb{S}_{23}$ &$\mathbb{S}_{31}$&$\mathbb{S}_{12}$\\ \hline
     Left Wheel&2&3&1\\ \hline
     Right wheel&3&1&2\\ \hline
     Normal vector&$e_{n1}$&$e_{n2}$&$e_{n3}$\\ \hline
     Tracking error&$e_{n1}$&$e_{n2}$&$e_{n3}$\\ \hline
\end{tabularx}
\end{table}
For the three-wheeled system with six \ac{dof}s, similar to the hybrid motion/force control in Sec.~\ref{sec:control}, the force along $\bm{z}_B$ and the linear motion along $\bm{x}_B$ and $\bm{y}_B$ are feedback controlled. The torque control related to the angular dynamics of the system however needs further investigation, since there are two more \ac{dof}s compared to the planar case. \new{We define the torque control wrench along $\bm{x}_B$ and $\bm{y}_B$ used for tip-over recovery and avoidance as $\bm{\tau}_2 \in \mathbb{R}^2$.} The platform orientation along $\bm{z}_B$ does not contribute to the tip-over behavior, therefore the control wrench along $\bm{z}_B$ is obtained via attitude control. The three wheels form a support polygon with three sides as in Fig.~\ref{fig:omni}. Each side of the support polygon is defined by two contact points which can be interpreted as a two-wheeled planar system. Each plane of the interpreted planar system from all three sides of the support polygon is defined in Table~\ref{table:planes}. The torque control wrench $\bm{\tau}_2$ is thus the sum of the torque control wrench of each plane in 3D space. To combine the torque control wrench of each plane in body frame, we define three unit vectors $\bm{n_1}$, $\bm{n_2}$, $\bm{n_3} \in \mathbb{R}^3$ parallel to the plane $(\bm{x}_B, \bm{y}_B)$ indicating the positive direction of the rotation and torque in planes $\mathbb{S}_{23}$, $\mathbb{S}_{31}$ and $\mathbb{S}_{12}$, respectively, see Fig.~\ref{fig:omni}. Knowing the force measurements on three wheels $f_{n1}$, $f_{n2}$ and $f_{n3}$ respectively, based on the tip-over stability criterion in Sec.~\ref{sec:tip-over}, the full-contact scenario of the three-wheeled system is defined by: $f_{ni}\geq \delta_F, \text{with $i=1,2,3$}$.
Similarly, the nominal contact case is given by:
\begin{equation}
    f_{n1}=f_{n2}=f_{n3}.
\end{equation}
The corresponding force difference errors of each plane are defined as $e_{n1}=f_{n3}-f_{n2}$, $e_{n2}=f_{n1}-f_{n3}$, and  $e_{n3}=f_{n2}-f_{n1}$.
The torque control wrench $\bm{\tau}_{2}$ is given by:
\begin{equation}
\bm{\tau}_2=k_{n1}e_{n1}\bm{n}_1+k_{n2}e_{n2}\bm{n}_2+k_{n3}e_{n3}\bm{n}_3,
\end{equation}
where $k_{n1}$, $k_{n2}$, $k_{n3}$ are positive gains. This methodology can also be applied to four-wheeled systems or other multi-wheeled layouts.

\section{Validation}\label{sec:validation}
In this section, we present the validation of the high-level control design in Sec.\ref{sec:control} and benchmark the mechanical design guidelines in Sec.~\ref{sec:guidelines} under measured systematic uncertainties. To test the robustness of the proposed approaches in risky scenarios, e.g., sliding maneuvers and small pushing force, we use simulations to avoid hazardous operations of the real platform. Moreover, to take into account the uncertainties from the real low-level system, we identify the force and torque uncertainties at the \ac{com} of the system using FT sensor measurements from physical experiments. Later, a simulator based on the system in Fig.~\ref{fig:force} is built to simulate the tip-over problem for passive sliding with fully-actuated aerial vehicles. We calibrate the simulator based on the experimentally determined uncertainties in force-torque application and end-effector pressure sensing. The measured uncertainties take into account the modeling errors from force estimation, state estimation, low-level control, and sensor noises. 

The system performance is evaluated by the force difference error $e_n$ \eqref{eq:force_error} from contact points and the contact condition $\beta$ in Table~\ref{table:contact} based on Sec.~\ref{sec:tip-over}. In total, four scenarios are tested for tip-over avoidance, involving different sliding accelerations and pushing force magnitudes. As a baseline, we compare against the simulation results using impedance control with \new{ the same wheel distance $r_d=\SI{0.084}{\meter}$ as the real platform}, which reflects the physical experiments in Sec.~\ref{sec:case_study}. Following the guidelines in Sec.~\ref{sec:guidelines}, a reasonable hardware design to avoid tip-over considering the platform size is to have twice the wheel distance as $2r_d$, based on the results in Fig.~\ref{fig:a_up}. Two approaches including using an enlarged end-effector ($2r_d$) with the baseline impedance control and an implementation of the control design in Sec.~\ref{sec:tipover_control} are tested and compared. In a high-risk scenario, we evaluate the impact of variable contact angles (i.e., the relative angles between the surface and the end-effector tip) to validate the robustness of the normal-force controller in tip-over recovery. The control design in Sec.~\ref{sec:3d} for a three-wheeled system is also validated with a simulator calibrated with measured uncertainties and modeled pressure sensor noises in high-risk scenarios.

\subsection{Uncertainty Identification}
\begin{figure}[!t]
\centering
\includegraphics[width=\columnwidth]{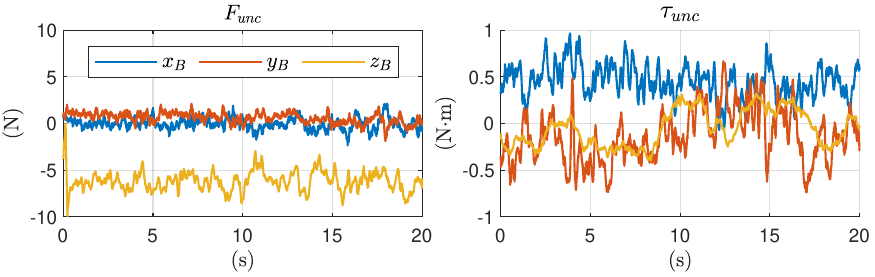}
\caption{Measured force and torque uncertainties at the \ac{com} of the system from the low-level system expressed in body frame.}
\label{fig:dis}
\end{figure}

In physical experiments, we used the FT sensor shown in Fig.~\ref{fig:wheel_num} to measure the external forces $\bm{F}_{meas} \in \mathbb{R}^3$ and torques $\bm{\tau}_{meas} \in \mathbb{R}^3$ acting on the mounting point of the sensor in body frame when the system is in contact. We denote $\bm{p}_{ft} \in \mathbb{R}^3$ as the position vector from the mounting point of the FT sensor to the \ac{com} of the system expressed in the body frame. With the parameters in Sec.~\ref{sec:platform}, we have $\bm{p}_{ft}=\begin{bmatrix}
    0&0&-h_{ft}
\end{bmatrix}^{\top}$ for the hardware setup. During the experiments, the desired pushing force along $\bm{z}_B$ was \SI{10}{\newton}. With the rigid body assumption, the linear forces acting on the \ac{com} of the system are equal to $\bm{F}_{meas}$. The torques acting on the \ac{com} are calculated via a simple transformation: $\bm{\tau}_{meas}^{CoM}=\bm{\tau}_{meas}+\bm{F}_{meas} \times \bm{p}_{ft}$.
From the model-based force estimation in the low-level system, we obtain the estimated external forces and torques at the \ac{com} as $\bm{F}_{est}$ and $\bm{\tau}_{est}$ expressed in body frame. With the above information, we identify systematic uncertainties in the form of force and torque generation expressed in body frame by: $\bm{F}_{unc}=\bm{F}_{meas}-\bm{F}_{est}$, and $\bm{\tau}_{unc}=\bm{\tau}_{meas}^{CoM}-\bm{\tau}_{est}$.
The experimentally obtained force and torque uncertainties at the \ac{com} are displayed in Fig.~\ref{fig:dis}. The results show that there are significant modeling errors in force estimation since the force error magnitude along axis $\bm{z}_B$ is almost the same as the desired force value. The uncertainties in angular dynamics $\bm{\tau}_{unc}$ hint at \ac{com} offsets, which are crucial in push-and-slide tasks. The identified uncertainties are later integrated with the actuation wrenches from the high-level controller in the simulator.
\subsection{Simulator Setup}
 A Simscape model based on the system described in Fig.~\ref{fig:force} is used for testing. The proposed controller in Sec.~\ref{sec:control} and the baseline controller are implemented via Matlab/Simulink at 100Hz. The Simscape model captures the main geometric parameters of the real platform, i.e., the wheel radius and wheel distance. The friction coefficients at contact are set to have similar values to the real experiment setup during passive sliding tasks. In the simulation, we command the system to approach a flat work surface with different contact angles $\delta\beta$ to simulate the tip-over at initial contact with an unknown surface. Once the system gains full-contact, it will slide along the work surface with a desired maximum acceleration $a_{max}=$\SI{0.5}{\metre\per\square\second}. To capture the effects of uncertainties with higher magnitude, the identified force uncertainties along $\bm{x}_B$ and $\bm{z}_B$, and the identified torque uncertainties along axis $\bm{x}_B$ are integrated with $f_{X_B}$, $f_{Z_B}$ and $\tau_B$ from the high-level controller, respectively. The data in Fig.~\ref{fig:dis} is resampled according to the simulation sample time to be simultaneous with the actuation wrenches in the simulator. The simulated actuation wrenches are given by: $f_{X_B}^{sim}=f_{X_B}-\begin{bmatrix}
        1&0&0
    \end{bmatrix}\bm{F}_{unc}$, $ f_{Z_B}^{sim}=f_{Z_B}-\begin{bmatrix}
        0&0&1
    \end{bmatrix}\bm{F}_{unc}$, $\tau_B^{sim}=\tau_B-\begin{bmatrix}
        1&0&0
    \end{bmatrix}\bm{\tau}_{unc}$.
 Moreover, the pressure sensor noises are simulated via a Gaussian distribution with a scale of $10$ percentage of the data magnitude, which is a choice based on preliminary tests with pressure sensors. With the simulator setup, we can test the controller behavior for both tip-over recovery and avoidance in the presence of uncertainties. We apply a similar approach to include the measured uncertainties in the tests using the baseline impedance controller.

The control gains related to motion and force control used in \eqref{eq:motion} and \eqref{eq:force} are selected according to the baseline impedance controller used in the previous work. For the gain $k_n$ used in the normal-force-based torque control in \eqref{eq:torque}, a heuristic tuning method was applied. An initial guess was based on the ratio between the magnitude of the system's maximum applicable torque within saturation and the desired pushing force $f^d$. Then we adjusted the initial gain value until the system gave a robust tip-over recovery and avoidance behavior even in the presence of the measured uncertainties. For simulating the actuator saturation of the physical platform, the upper and lower limits of the controlled torque are set to $\pm 5$ \si{\newton\meter} in the simulations, with $k_n=0.5$. The margin $\delta_F$ introduced in Sec.~\ref{sec:tip-over} is set to $1\si{\newton}$.
\subsection{Robustness Test for Tip-Over Avoidance}
\begin{figure}[!t]
\centering
\includegraphics[width=\columnwidth]{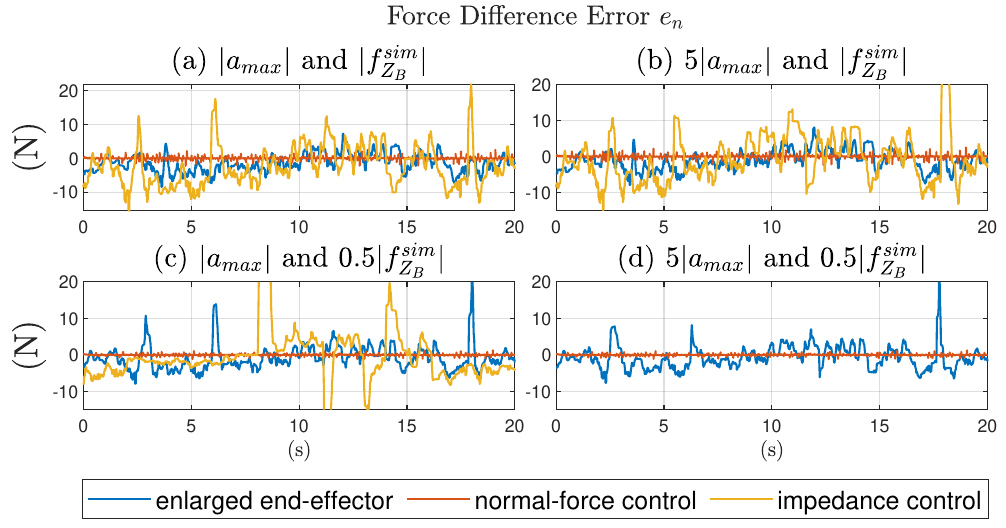}
\caption{Force difference error $e_n$ for differently scaled $|a_{max}|$ and $|f_{Z_B}^{sim}|$ benchmarking the impedance control, enlarged support polygon with impedance control, and the normal-force control. In (d) no data is displayed for the impedance control case due to instability at initial contact.}
\label{fig:robust}
\end{figure}

\begin{figure}[!t]
\centering
\includegraphics[width=\columnwidth]{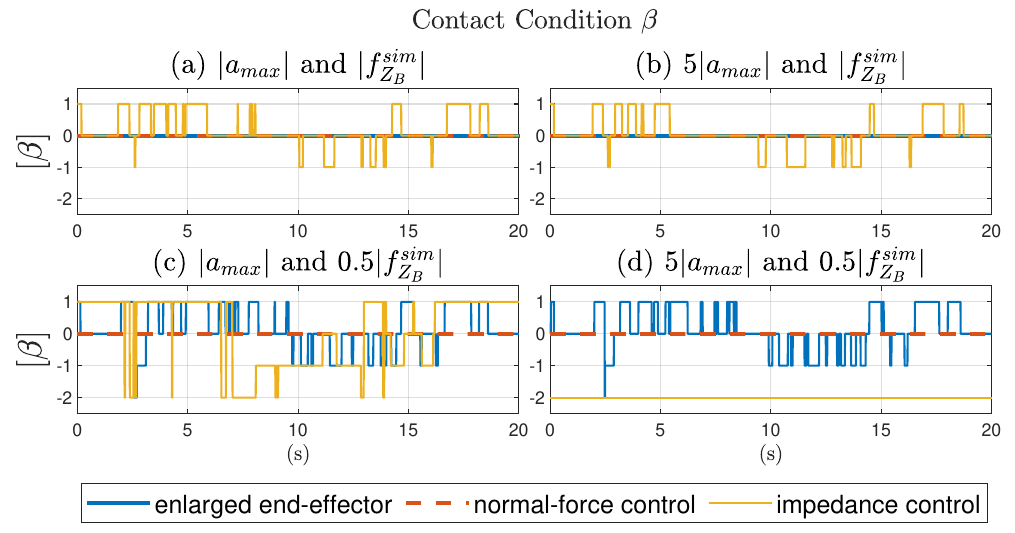}
\caption{Contact conditions: for impedance control tip-overs occurred in all scenarios, for enlarged end-effector tip-overs occurred in (c) and (d), and for normal-force control, no tip-overs occurred in all scenarios. In (d) instability happened for the impedance control case at initial contact, and $\beta$ is presented by $-2$ same as the value for free flight. }
\label{fig:beta}
\end{figure}

\begin{table}[!t]
\caption{Testing Scenarios}\label{table:scenarios}
\begin{center}      
    \begin{tabularx}{\columnwidth}{p{0.05\textwidth}|p{0.18\textwidth}|X} 
     \hline
    Case Num.& maximum sliding acc (\SI{}{\metre\per\square\second}) & pushing force (\SI{}{\newton})\\ 
     \hline
     (a)  & $|a_{max}|$&$|f_{Z_B}^{sim}|$\\
     \hline
     (b)  & $5|a_{max}|$&$|f_{Z_B}^{sim}|$\\ \hline
     (c)  & $|a_{max}|$&$0.5|f_{Z_B}^{sim}|$\\ \hline
     (d)  & $5|a_{max}|$&$0.5|f_{Z_B}^{sim}|$\\ \hline
\end{tabularx}
\end{center}
\end{table}
In this section, we present the robustness tests using differently scaled maximum sliding acceleration $|a_{max}|=$\SI{0.5}{\metre\per\square\second} and pushing force $|f_{Z_B}^{sim}|\approx$\SI{15}{\newton} for tip-over avoidance during sliding assuming that the initial alignment is achieved. To do so, four scenarios with the simulation setup in the last section are tested as shown in Table.~\ref{table:scenarios}. Based on the analysis in Sec.~\ref{sec:force_angle}, the testing scenario is riskier when larger sliding accelerations or smaller pushing forces are used. Therefore (d) is the most critical testing scenario among the four. As a comparison, the baseline impedance controller, the enlarged end-effector with impedance control, and the control scheme in Sec.~\ref{sec:control} are tested in all four scenarios. The resulting force difference error $e_n$ and the contact condition $\beta$ are shown in Fig.~\ref{fig:robust} and Fig.~\ref{fig:beta} respectively. Fig.~\ref{fig:beta} indicates that tip-overs occurred for all testing scenarios when the baseline impedance control was applied. The $e_n$ resulting from using the impedance controller under scenario (d) is not shown in Fig.~\ref{fig:robust} (d) because instability occurred at initial contact. In Fig.~\ref{fig:beta}, the enlarged end-effector did avoid tip-over for the first two scenarios (a) and (b), however tip-over happened in riskier scenarios (c) and (d) when the pushing force is smaller. The normal-force controller instead effectively avoided tip-over for all four scenarios including sliding \new{maneuvers} and a smaller pushing force with very small errors $e_n$ as in Fig.~\ref{fig:robust}. Even in scenario (a) and (b) in Fig.~\ref{fig:robust}, when tip-over was avoided by both enlarging the end-effector and the normal-force control, the error $e_n$ from the normal-force control has a much smaller magnitude and oscillation compared to the one from the enlarged end-effector. From these results, we see that the enlargement of the end-effector is an effective approach to alleviate the impact of contact uncertainties on the system during sliding in low-risk scenarios. However, the proposed normal-force control approach is effective and robust even in high-risk scenarios.

\subsection{Robustness Test for Tip-Over Recovery}
In this section, we validate the robustness of the designed controller for tip-over recovery from initial misalignment under the most critical scenario from the last section: scenario (d) with $5|a_{max}|$ and $0.5|f_{Z_B}^{sim}|$. We let the system approach the work surface with a contact angle $\delta\beta$. With $\delta\beta=0$, we can simulate the possible misalignment due to systematic uncertainties. With $\delta\beta \neq 0$, we can simulate the case when the system approaches an unknown surface with a wrong attitude reference e.g., due to uncertainties in estimating the surface normal, where initial tip-over occurs. The tip-over recovery operation is tested for $\delta\beta= 0\degree, 10\degree, 20\degree$, with the later ones being very critical cases in reality. The resulting force difference error $e_n$ is shown in Fig.~\ref{fig:3w} (I), in which the high peak values indicate the initial impact when the wheel contacts the surface, which is quite high for larger contact angles. The system can recover from the initial tip-over for all three cases. From the results, we conclude that the normal-force control can robustly recover from tip-over situations.
%
\subsection{Three-Wheeled System}
\begin{figure}[!t]
\centering
\includegraphics[width=\columnwidth]{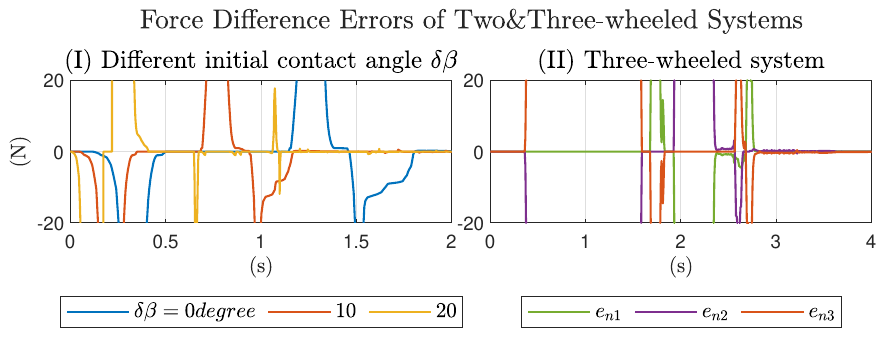}
\caption{Tip-over recovery and avoidance with normal-force control under $5|a_{max}|$ and $0.5|f_{Z_B}^{sim}|$ for (I): different initial contact angle $\delta\beta=0 \degree, \ 10 \degree, \ 20 \degree$ of the planar system; (II): contact angle $-5\degree$ in $\bm{y}_B$ and $15\degree$ in $\bm{x}_B$ of the three-wheeled system.}
\label{fig:3w}
\end{figure}

Finally, a similar simulation setup is used for a three-wheeled system as in Fig.~\ref{fig:wheel_num}. The identified uncertainties in all six directions in Fig.~\ref{fig:dis} are integrated with the actuation wrenches. To validate the extended control design in Sec.~\ref{sec:3d}, we let the system approach the surface with different orientations in $\bm{x}_B$ and $\bm{y}_B$ with $5|a_{max}|$ and $0.5|f_{Z_B}^{sim}|$. An example of approaching the surface with angle derivation w.r.t. the surface as $-5\degree$ in $\bm{y}_B$ and $15\degree$ in $\bm{x}_B$ is shown in Fig.~\ref{fig:3w} (II), where the force difference errors $e_{n1}$, $e_{n2}$ and $e_{n3}$ of three planes all converge to the equilibrium state. The system can gain full-contact of three wheels with the surface and avoid tip-overs in risky scenarios. With the results, we demonstrated the use case of the normal-force control design for three-wheeled systems applying the proposed method in Sec.~\ref{sec:3d}, which promises the generality of such a control strategy.

\section{Conclusion}\label{sec:conclusion}
In this work, we presented a comprehensive analysis and provided effective solutions to enhance sliding performance with aerial vehicles in non-actuated multi-wheel configurations. By conducting physical experiments with a fully-actuated aerial vehicle, we identified a common challenge in such tasks: maintaining stable contact between each wheel and the work surface. To address this issue, we applied related work from mobile manipulators to aerial systems, leading to the development of two approaches: one focused on hardware design guidelines and the other on a novel pressure-sensing-based interaction control framework. Both approaches were validated and evaluated using a realistic simulator grounded in experimental data and accounting for real-world uncertainties. Our analysis and proposed solutions not only resolve current challenges but also highlight promising avenues for future research in the use of multiple non-actuated wheels for sliding operations with aerial robots.

\section{Acknowledgement}
We thank Eugenio Cuniato, Michael Pantic, and Christian Lanegger from ETH for their help with conducting the experiments and valuable guidelines. 


\bibliographystyle{elsarticle-num-names}
\bibliography{references}








\end{document}